\documentclass[sigconf]{acmart}
\usepackage{colortbl}
\AtBeginDocument{%
  }

\copyrightyear{2026}
\acmYear{2026}
\setcopyright{cc}
\setcctype{by}
 \acmConference[DIS '26]{Designing Interactive Systems Conference}{June 13--17, 2026}{Singapore, Singapore}
\acmBooktitle{Designing Interactive Systems Conference (DIS '26), June 13--17, 2026, Singapore, Singapore}
\acmDOI{10.1145/3800645.3813000}
\acmISBN{979-8-4007-2563-0/2026/06}

\usepackage{xcolor,colortbl}

\usepackage{soul}
\usepackage[ruled,vlined]{algorithm2e}

\newcommand{\eg}[0]{e.g.,}
\newcommand{\ie}[0]{i.e.,}
\newcommand{\tool}[0]{\textit{Distill}}
\newcommand{\cmd}[2]{\texttt{\textbf{#1}}(\textit{#2})}
\newcommand{\struct}[0]{\textit{structured}}
\newcommand{\open}[0]{\textit{open-ended}}
\newcommand{\qt}[1]{\textit{``#1''}}

\newcommand{\pone}[0]{\textit{Phase 1 - NL}}
\newcommand{\ptwo}[0]{\textit{Phase 2 - Trace}}
\newcommand{\pthree}[0]{\textit{Phase 3 - Filter}}
\newcommand{\pfour}[0]{\textit{Phase 4 - Abstract}}
\newcommand{\pfive}[0]{\textit{Phase 5 - Group}}

\newcommand{\bl}[0]{\color{black}}
\newcommand{\bk}[0]{\color{black}}

\definecolor{rowgray}{gray}{0.93}

\definecolor{lred}{RGB}{0,0,0}

\begin{document}

\title{\textit{Distill}: Uncovering the True Intent behind Human-Robot Communication}

\author{Ting Li}
\affiliation{%
  \institution{Computer Science}
  \institution{George Mason University}
  \city{Fairfax}
  \state{Virginia}
  \country{USA}
}
\email{tli21@gmu.edu}

\author{David Porfirio}
\affiliation{%
  \institution{Computer Science}
  \institution{George Mason University}
  \city{Fairfax}
  \state{Virginia}
  \country{USA}
  }
\email{dporfiri@gmu.edu}

\renewcommand{\shortauthors}{Li \& Porfirio}

\begin{abstract}

As robots become increasingly integrated into everyday environments, intuitive communication paradigms such as natural language and end-user programming have become indispensable for specifying autonomous robot behavior. However, these mechanisms are ineffective at fully capturing user intent---natural language is imprecise and ambiguous, whereas end-user programming can be overly specific. As a result, understanding what users \textit{truly} mean when they interact with robots remains a central challenge for human-AI communication systems. To address this issue, we propose the \textit{Distill} approach for human-robot communication interfaces. Given a task specification provided by the user, \textit{Distill} (1) removes unnecessary steps; (2) generalizes the meaning behind individual steps; and (3) relaxes ordering constraints between steps. We implemented \textit{Distill} on a web interface, and through a crowdsourcing study, demonstrated its ability to elicit and refine user intent from initial task specifications.


\end{abstract}

\ccsdesc[500]{Human-centered computing~Interactive systems and tools}
\ccsdesc[300]{Computing methodologies~Robotic planning}
\ccsdesc[500]{Human-centered computing~HCI theory, concepts and models}

\keywords{Human–Robot Interaction, AI Planning, Intent Elicitation, Human-AI Communication, Interface Design }



\begin{teaserfigure}
    \includegraphics[width=\textwidth]{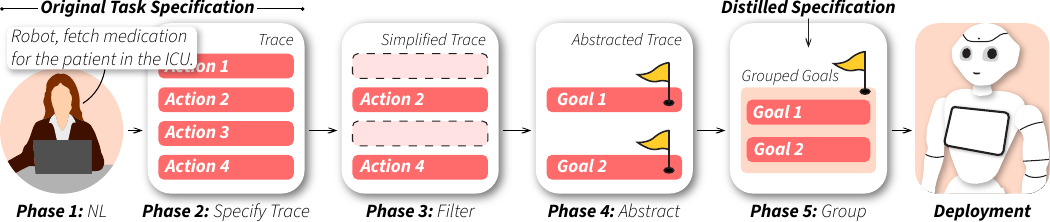}
    \caption{The \tool{} approach to eliciting \textit{ground-truth} user input from natural task specification paradigms.}
    \Description{Diagram of \tool{} system workflow showing five phases: user provides natural language commands, system generates an initial task trace, filters redundant steps, abstracts core actions, and groups goals for deployment, illustrating the transformation from input to structured executable plan.}
    \label{fig:teaser}
\end{teaserfigure}
\maketitle


\section{Introduction}\label{sec:intro}

As robots become ubiquitous in everyday life, end users increasingly rely on intuitive communication paradigms as convenient ways to direct their behavior. Most often, these paradigms fall into two categories---\textit{natural language} and \textit{end-user programming} (EUP). Natural language typically involves users issuing real-time commands of varying degrees of complexity to their robot teammates, such as \textit{``Deliver medication to the patient in room 1201. Make sure the patient in 1205 also gets their dinner.''} With EUP, the user can instead issue the same task by providing the robot with a hand-crafted sequence of symbolic instructions, typically via a graphical user interface: \cmd{moveTo}{pharmacy}, \cmd{request}{medication}, \cmd{moveTo}{1201}, \cmd{handoff}{medication, patient}, etc.

Although both methods have been shown to be exceptionally intuitive, they obscure the true intent of the user. Natural language, in particular, is imprecise and ambiguous \cite{piantadosi2012communicative}. With the ordering of deliveries in the example above, does the user care if the patient in room 1201 receives their medication before the patient in 1205 receives their dinner? 
EUP suffers from the opposite problem. By requiring users to hand-craft symbolic sequences of (potentially branching and looping) actions (\ie{} \textit{programs}), the resulting output is highly context-specific, and thus overly precise, unambiguous, and brittle to perturbations in context. 
With the sequence of step-by-step actions above, does the user specifically intend for the robot to receive the medication from the pharmacy, or do they care if a nurse hands it to the robot instead? While the imperfections of both intuitive communication methods are not inherently unworkable---the robot can be trained to infer \bl{}human intent\bk{} from patterns within these imperfections that arise in specific contexts---the human-robot interaction (HRI) community lacks methods, tools, and datasets that extract the intended meaning behind imperfect user input.

Facilitating this extraction is crucial to enabling robots to infer \bl{}\textit{ground-truth} intent from imperfect human input. We define ground truth as an expression of intent that is both sufficient (contains \textit{all} information that is necessary) and minimal (\ie{} contains \textit{no more} information than is necessary) at capturing the user's task constraints and preferences.\bk{} A necessary first step is to design a pipeline, ideally operationalizable through any human-robot communication interface, that guides users to clarify their ground-truth intent after initially expressing robot tasks. A critical aim of this paper is thus to provide guidelines for constructing such a pipeline. Additionally, we believe that if future interfaces are both effective and efficient at extracting a user's ground-truth task intent, this may reduce the need for the robot to rely purely on inference. 


We thereby designed a prototype interface that helps uncover and clarify the intentions behind intuitive task specifications. The system first invites users to provide a natural language task description, followed by step-by-step instructions for how the robot should complete its task. The system then automatically \textit{distills} this plan to only include \textit{critical} actions \cite{porfirio2025bootstrapping}, namely by removing actions that are implied by future actions (\eg{} \cmd{moveTo}{medicine} is implied by \cmd{grab}{medicine}), while giving users the option to reintroduce any non-critical steps that they deem critical. The system then assists users in reflecting on the purpose behind each action and allows them to reorder or adjust actions when there are multiple high-level goals. Through this structured process, users articulate not only what they want the robot to do, but also reflect on how and why particular actions matter to them.

\bl{}Overall, we intend for this approach, which we call \tool{}, to\bk{} guide users to translate their initial natural language and step-by-step task specifications into a minimal, partially ordered sequence of critical steps. Beyond supporting ground-truth extraction, our work offers insights into how humans communicate with robots, revealing their goals, preferences, and expectations in open-ended tasks. \bl{}Our specific research questions are: (\textbf{RQ1}) which aspects of the \tool{} approach are effective, or ineffective, at eliciting ground-truth user intent; and (\textbf{RQ2}) how can \tool{} inform the design of future human-robot task communication interfaces?\bk{} This paper makes three key contributions in pursuit of answers to these questions, in increasing order of importance:

\begin{itemize}
    \item \textit{Design}---An approach to guiding users to translate their own natural input into a minimal, partially ordered set of \textit{critical} actions, clarifying what they truly mean.
    \item \textit{Empirical}---A user study that demonstrates the effectiveness of \tool{} at eliciting true user intent from initially provided natural input.
    \item \textit{Implication}---A set of design implications, based on the results from the study, for how interfaces should be constructed to elicit user intent.
\end{itemize}



\section{Related Work}
In order to situate our work within the broad space of human-robot task communication, we begin by discussing related work in natural language communication and end-user programming in human-robot interaction. We then provide a background on \textit{automated task planning}, which is key to understanding the research gap that our work addresses.

\subsection{Human-Robot Natural Language Communication}
\bl{}Alongside collaboration and cognition, \textit{communication} is regarded as one of the three pillars of human-robot teaming \cite{singh2026human}. A subset of human-robot communication involves natural language dialog\bk{}, in which the human issues commands to the robot \cite{shridhar2020alfred, porfirio2023crowdsourcing}, the robot requests clarification, commentary, or feedback \cite{lukin2024scout, padmakumar2022teach, dougan2022asking}, and in some cases, both entities engage in small talk \cite{pineda2025see, ramnauth2025robot}. This paper focuses specifically on communication that might occur at the start of \textit{pooled-interdependent} teaming \cite{zhao2020task}, in which the human provides a full, one-shot task specification to a robot that has high \textit{operational autonomy} \cite{kim2024taxonomy} at runtime with limited opportunity for human correction. This requires that the user communicate their intent to the robot \textit{up front} and \textit{in full}. Recently, there has been substantial prior work in large language models (LLMs) \cite{pmlr-v235-kambhampati24a, lee2025veriplan,porfirio2025bootstrapping} and vision-language action (VLA) models \cite{brohan2023can, pmlr-v229-zitkovich23a, quartey2025verifiably} that enable humans to express full tasks to the robot through natural language.

While these tools have proven extraordinarily useful, natural language suffers the limitation of being both imprecise and ambiguous \cite{piantadosi2012communicative}. \bl{}Compounding this problem, humans tend to communicate with robots differently than they do when communicating with other humans \cite{lukin2024scout}. For example, prior work shows that users tend to express commands \textit{indirectly} to the robot \cite{briggs2017enabling}. Furthermore, humans are more \textit{incremental} when communicating with robots, reasoning about the simple physical actions for the robot to take rather than higher-level belief updates \cite{marge2020let}. With such incrementality, task specifications can become fragile, as illustrated by \citet{liao2019synthesizing}, who demonstrate that the \textit{essential} steps of a task are hidden within users' sequence of instructions. In contrast to the essential steps, the original sequence is often environment-dependent, that is, not generalizing to new environments. Fortunately, robots can be trained to infer ground-truth user intent from free-form natural language queries \cite{misra2016tell} or resolve such ambiguities through interactive dialog \cite{chisari2025iros}. Insufficient inference, however, can decrease the user's perception of the robot and harm task performance \cite{zhang2025can, williams2018thank}. Rather than relying solely on the robot to recover intent, we instead ask: can we guide users to articulate a clearer representation of their ground-truth intent themselves?\bk{}


%


\subsection{Robot End-User Programming}

Broadly, end-user programming (EUP) enables non-roboticists to create robot applications \cite{ajaykumar2021survey}. These applications include task specifications \cite{huang2020vipo, cao2019ghostar, hagenow2024system, paxton2017costar}, namely the instructions for what a robot should do to achieve a set of task objectives, and social interactions \cite{gorostiza2011end, kubota2020jessie, glas2012interaction, cruz2025poder}, namely the behaviors that a robot should perform to be sociable. EUP is often \textit{symbolic} in nature, in which end-user programmers (henceforth referred to as \textit{users}) assemble the robot's desired control flow in terms of discrete actions (\eg{} \cmd{moveTo}{kitchen}) rather than continuous movements (\eg{} the continuous motions that the robot should make to move to the kitchen).

EUP tools can be categorized based on the programming paradigm that they support. Many support \textit{flow-based} paradigms, such as graphical block-based programming tools that enable users to assemble and nest instructions within conditionals and loops \cite{schoen2022coframe, schoen2023lively, schoen2024openvp, huang2016design, huang2017code3, zhang2025balancing}, or graphical state-based tools which enable users to construct programs represented as finite state automata \cite{pot2009choregraphe, alexandrova2015roboflow, porfirio2018authoring, cao2019v, porfirio2024goal}. A subset of these tools restricts users to producing linear sequences of steps on a timeline with no opportunity for branching or looping, which we refer to as \textit{traces} (short for \textit{execution traces}) \cite{sauppe2014design, schoen2020authr, porfirio2023crowdsourcing, porfirio2025bootstrapping}. In an age of increasing robot autonomy, hand-crafted traces are interesting sources of investigation. Traces with less detail---\eg{} \cmd{grab}{mop} then \cmd{wipe}{floor}---afford the robot greater autonomy to complete its task how it sees fit. These traces have been referred to as ``sketches'' in prior work \cite{liao2019synthesizing}, in which there are ``holes in place of low-level details.'' In contrast, traces with more detail---\eg{} \cmd{moveTo}{closet}, \cmd{open}{door}, \cmd{grab}{mop}, \cmd{close}{door}, \cmd{moveTo}{spill}, and \cmd{wipe}{floor}---further restrict how the robot performs its task \cite{porfirio2024goal}.

In the next section, we discuss automated task planning. Whereas EUP facilitates manual trace construction, automated task planning enables the robot to construct step-by-step actions by itself.

\subsection{Automated Task Planning}
Given a task specification---either a command expressed in natural language or a program constructed via an EUP tool---\textit{automated task planning} (henceforth referred to simply as \textit{planning}) is a mechanism behind which a robot can autonomously decide how to execute the task. Classically, planning problems consist of three components: (1) a set of actions that the robot can perform; (2) a set of entities in the robot's environment and the relations between them; and (3) a goal \cite{ghallab2016automated}. If it knows these three components, a robot can assemble a \textit{plan} to achieve the goal, consisting of the step-by-step actions that the robot must take. Several off-the-shelf libraries and algorithms have existed for decades for representing \cite{fox2003pddl2, smith2008anml} and solving \cite{hart1968formal, helmert2006fast, speck2023symk, micheli2025unified} planning problems.

With the explosion of generative artificial intelligence in recent years, large language models (LLMs) have shown to be useful for facilitating the direct translation from user natural language to step-by-step plans \cite{lee2025veriplan, pmlr-v235-kambhampati24a, shojaee2025illusion}, with VLAs and vision language models (VLMs) improving these capabilities by combining the user's natural language with entities captured by the robot's sensors \cite{brohan2023can, pmlr-v229-zitkovich23a, quartey2025verifiably}. For example, if the user asks the robot to ``use the mop to clean the spill'' and the robot observes liquid on the ground and cleaning supplies nearby, the robot will know to (1) travel to the closet, (2) open the door, (3) pick up the mop, (4) close the door, (5) travel to the spill, and (6) mop up the liquid.

A key concern with natural language is its ambiguity \cite{piantadosi2012communicative}, which obfuscates user preferences. Suppose that the user instead says, ``clean the spill,'' and there are several different cleaning supplies nearby. 
Should the user have an implicit preference for what the robot should use, natural language lacks information that could be useful to the robot.

EUP offers a solution to this ambiguity by affording users direct plan interaction. Rather than inadvertently omitting critical information in natural language, recent advances in EUP have enabled users to hand-craft task specifications with an appropriate level of detail for guiding the planner while maintaining adherence to user preferences \cite{porfirio2024goal}. For example, users who care simply that the spill is clean can hand-craft a single goal---\cmd{isClean}{floor} (translates to ``the floor is clean'')---and let the robot figure out how to achieve it. Other users might prefer that the robot wipes the spill with a mop rather than a rag, and thus add specificity by providing the robot with the preferred action---\cmd{mop}{floor}. Others may wish to add further specificity by requesting that the robot grab the mop from the laundry room rather than the closet---first, \cmd{moveTo}{laundryRoom}, then \cmd{grab}{mop}, and finally \cmd{mop}{floor}.

Prior work in EUP shows that users err towards higher specificity \cite{porfirio2023crowdsourcing, porfirio2025interaction}, often to the robot's detriment \cite{porfirio2024goal}. If the environment changes at all---for instance, if the mop is moved from the laundry room---the robot will need to infer which actions it can change from the user's original specification. With EUP, users naturally provide too much information to the robot, from which the robot must extract the user's core meaning.



\section{The \tool{} Approach}\label{sec:approach}

We present \tool{}, a five-phase interaction approach designed to help users communicate task specifications to autonomous robots. Rather than requiring users to learn formal specification languages or provide complete executable plans, \tool{} guides users to progressively refine natural language instructions into structured representations that capture user intent at multiple levels of abstraction. Figure~\ref{fig:teaser} illustrates the overall workflow, showing how each phase transforms the user's input from an original task specification toward a distilled specification suitable for robot deployment.
 
Each phase builds on the previous one, progressively revealing aspects of the user's ground-truth intent from (1) natural language and (2) procedural traces to (3) filtered, (4) abstracted, and (5) grouped task specifications. We provide implementation-agnostic details for each phase below, reflecting on different ways in which each phase can be operationalized in different interfaces. Following our implementation-agnostic descriptions, we describe our own implementation of \tool{} on a web interface. Note that the primary purpose of our web interface is to serve as a research platform for our evaluation of \tool{}, and should thus not be considered a standalone tool.

\bl{}To help illustrate how user input can be misinterpreted without \tool{}, consider the first part of the task introduced in Section \S\ref{sec:intro}, in which the robot (either verbally or through EUP) is instructed to get the medication from the pharmacy, move to room 1201, and then hand off the medication to the patient. Suppose that while traveling to room 1201, the robot encounters the patient in the hallway. Despite knowing the patient's true whereabouts, a naïve robot will still travel to room 1201 to adhere to the nurse's original instructions, only to then backtrack, travel to where it knows the patient is, and then deliver the medication to them. The plan includes an unnecessary step of traveling to room 1201, adding time and effort to the task. Distill prevents this by extracting the nurse's underlying intent, recognizing that the goal is simply to retrieve the medication from the pharmacy and deliver it to the patient \textit{regardless} of their current location, and generating a trace that goes directly to the patient without the redundant intermediate step.\bk{}

\subsection{Phases 1 and 2: Initial Task Expression}

\begin{figure}[!h]
    \includegraphics[width=\columnwidth]{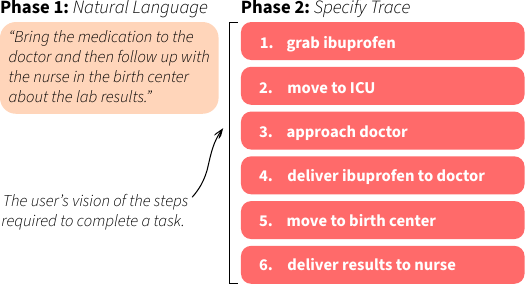}
    \caption{Example input to the first and second phases of the \tool{} approach.}
    \Description{Two-panel schematic showing Phase 1 natural language task input and Phase 2 corresponding initial task trace as a sequence of user-created states or actions.}
    \label{fig:12abs}
\end{figure}

Phases 1 and 2 ask that the user provide task input to the robot in \textit{natural} ways---specifically, through how they would normally envision themselves communicating with a robot---either through natural language and/or through a task trace. These phases are illustrated in Figure \ref{fig:12abs}.

\paragraph{Phase 1: Natural Language (NL)} Users describe a task for an autonomous robot using free-form natural language. We envision this natural language being provided either via typing on a keyboard or touchpad or via spoken language. In our hospital scenario, for example, another command from the nurse might be, \qt{Bring the medication to the doctor and then follow up with the nurse in the birth center about the lab results.} Note that the phrase \qt{and then} implies task sequentiality, even though the nurse might not actually care which order the deliveries are made. Furthermore, the phrase \qt{follow up with} is ambiguous, obfuscating the nurse's true intent.

\paragraph{Phase 2: Specify Trace}
This phase mirrors common approaches to end-user programming \citep[\eg{}][]{huang2016design, huang2017code3, schoen2023lively, schoen2024openvp, alexandrova2015roboflow, cruz2025poder, kubota2020jessie, porfirio2023crowdsourcing}. Specifically, users construct a step-by-step task trace by selecting and parameterizing actions from a predefined toolbox of domain-specific primitives. In our hospital example, these actions might be \cmd{grab}{ibuprofen}, \cmd{moveTo}{ICU}, \cmd{approach}{doctor}, etc. Note that each action requires appropriate parameters (\eg{} target objects, locations, recipients). Overall, this phase externalizes users' procedural knowledge, revealing their assumptions about required actions, causal dependencies, and execution order. Essentially, traces are meant to be a more concrete demonstration of \textit{how} users believe the task should be executed, which natural language struggles to capture. \bl{} It is important to clarify that \tool{} does not encourage users to over-specify their traces, nor does it consider over-specification a goal of this phase. Instead, over-specification is a natural and well-documented tendency in hand-crafted step-by-step task instructions \cite{porfirio2025interaction, porfirio2023crowdsourcing}. \tool{} accommodates this tendency by allowing users to specify tasks as they naturally would, while relying on subsequent filtering and abstraction phases to identify and remove redundant detail.\bk{}

\paragraph{Design Addendum. } In practice, the first two phases may vary depending on how \tool{} is operationalized across different user interfaces. In some user interfaces, users may prefer to provide spoken-language input to the robot, from which an LLM can automatically produce a trace as demonstrated by prior work \cite{pmlr-v235-kambhampati24a, lee2025veriplan, porfirio2025bootstrapping}. The user can then check and edit the trace to ensure it is accurate. Alternatively, perhaps the user may prefer to hand-craft a trace, skipping the first phase altogether. This is also fine. The main objective of these first two phases is to produce step-by-step instructions for the robot to execute, regardless of how these instructions are created. Our emphasis on step-by-step instructions is motivated by prior work showing that such instructions are intuitive for users to produce and comprehend \cite{porfirio2023crowdsourcing, porfirio2024goal, porfirio2025interaction, trafton1991providing}.

\subsection{Phase 3: Filter}
In this phase, \tool{} \textit{filters} participant traces. This can intuitively be understood as ``reverse planning.'' In contrast to \textit{planning}, which constructs sequences of actions for the robot to perform given a small set of goals, \textit{filtering} instead removes actions from user traces that are redundant. Specifically, it analyzes the user-created task trace to identify a minimal set of \textit{critical} actions that the user could have expressed to the robot for it to achieve the same result. Actions inferred to be redundant or inferable are filtered out as \textit{non-critical}. The distinction between critical and non-critical is made based on the robot's autonomous task planning abilities. In autonomous task planning, telling the robot to take action $A$ may require it to execute a sequence of preceding actions $\alpha_1, \alpha_2, ..., \alpha_n$ first. Using the hospital scenario for illustration in Figure \ref{fig:12abs}, the robot knows that in order to execute the action \cmd{deliver}{ibuprofen, doctor}, it must first grab the ibuprofen and find the doctor in the ICU. Grabbing the ibuprofen and finding the doctor would be deemed as non-critical in this case, because they are implied by the delivery action.

\begin{figure*}[!h]
    \centering
    \includegraphics[width=\textwidth]{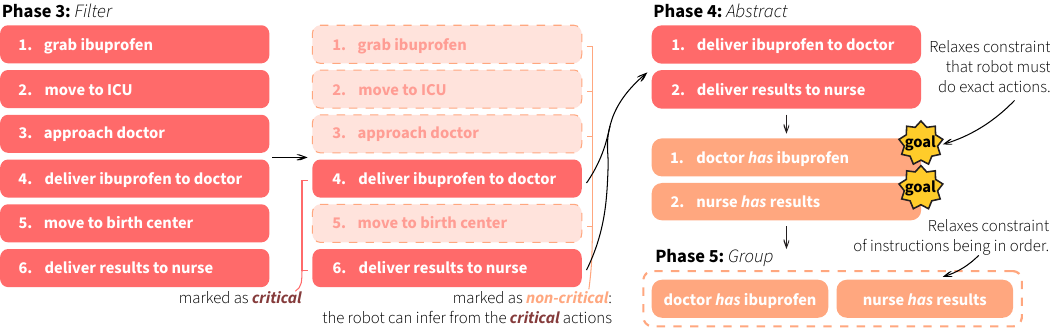}
    \caption{\tool{}'s third phase (left) involves filtering non-critical actions from the user's initial task trace. \tool{}'s fourth phase (top-right) relaxes the constraint that the robot must perform specific actions in order to achieve desired goal predicates. The fifth phase (bottom-right) relaxes the constraint that the robot must follow instructions in a certain order.}
    \Description{Combined figure of Distill Phases 3–5: Phase 3 removes non-critical actions from a task trace; Phase 4 abstracts actions into goals; Phase 5 groups goals to relax ordering constraints and support more flexible execution.}
    \label{fig:345abs}
\end{figure*}

Prior work shows that users intuitively insert non-critical actions in the steps that they provide to the robot \cite{porfirio2025interaction}, and in the spirit of supporting intuitive task specification, the previous phase of \tool{} allows for this to occur. Non-critical actions can mislead the robot, though. To illustrate this point, consider an example in which the doctor is not in the ICU. Telling the robot to move to the ICU before making the delivery may cause the robot to unnecessarily move to the wrong area of the hospital in its quest to find the doctor. Thus, the sole purpose of this phase is to remove non-critical actions from the user's trace and to provide the robot with the necessary and sufficient steps that it needs to complete its task. Figure \ref{fig:345abs} illustrates this process using our hospital scenario as an example.

There are several different ways that the removal of non-critical actions can be done, ranging from using LLMs to using deterministic ``reverse planning'' algorithms. We leave the description of our implementation to Section \S\ref{sec:implementation}. Crucially, we advocate that users maintain the ability to review the filtered trace and can override \tool{}'s classification of critical and non-critical actions. Filtering surfaces mismatches between algorithmic and human judgments of essentiality, revealing which details users expect to specify versus delegate to the robot. Allowing user validation supports mixed-initiative refinement \cite{horvitz1999principles} of the filtering process.

\subsection{Phase 4: Abstract}
In this phase, for each action in the filtered trace, users are prompted to indicate whether they care about the robot performing that \textit{specific action} or only about achieving the \textit{postconditions} of that action (the \textit{goals}). Outcomes are represented as goal predicates. In the hospital example illustrated in Figure \ref{fig:345abs} (top right), an outcome of \cmd{deliver}{ibuprofen, doctor} would be that \textit{the doctor has ibuprofen}.

Prior work shows that users intuitively think about tasks in terms of step-by-step actions rather than outcomes \cite{porfirio2024goal, trafton1991providing}, and in the spirit of supporting intuitive task specification, the previous phases of \tool{} facilitate action-level task specification. Outcome-oriented specification, however, further enables the robot to figure out how to achieve the true task intent on its own terms. For example, there are multiple ways in which the robot can achieve the outcome of the doctor having ibuprofen---the robot can request that a nurse deliver it if the doctor is unavailable, for example. Therefore, the purpose of this phase is to disentangle actions from desired effects, leveraging flexibility in how goals may be achieved. By separating what must be accomplished from how it is done, \tool{} thereby supports more robust planning under environmental uncertainty.

\subsection{Phase 5: Group}
In this phase, users are prompted to indicate whether the order of certain steps can be relaxed. In the hospital example illustrated in Figure \ref{fig:345abs} (bottom right), the user might indicate that they do not care which outcome the robot achieves first---the doctor can receive ibuprofen either before or after the nurse receives the lab results. Users thereby specify temporal preferences by grouping goals and assigning priorities. Goals within the same priority group may be executed in any order or in parallel, while different priority groups impose sequencing constraints (\eg{} the group with priority of ``1'' must complete before the group with priority ``2'' begins). Note that Figure \ref{fig:345abs} (bottom right) only indicates a single group, whereas in practice, there may be many more steps organized into several different groups of varying priorities.

The purpose of this phase is to capture users' preferences about execution ordering and parallelization opportunities, which is an often omitted feature of state-of-the-art end-user programming tools that focus solely on sequential execution. By explicitly representing temporal flexibility, \tool{} enables more efficient robot execution while respecting user constraints. The grouping mechanism supports both strict sequencing (separate priority groups) and flexible ordering (actions within the same group).

\begin{figure*}
    \centering
    \includegraphics[width=\textwidth]{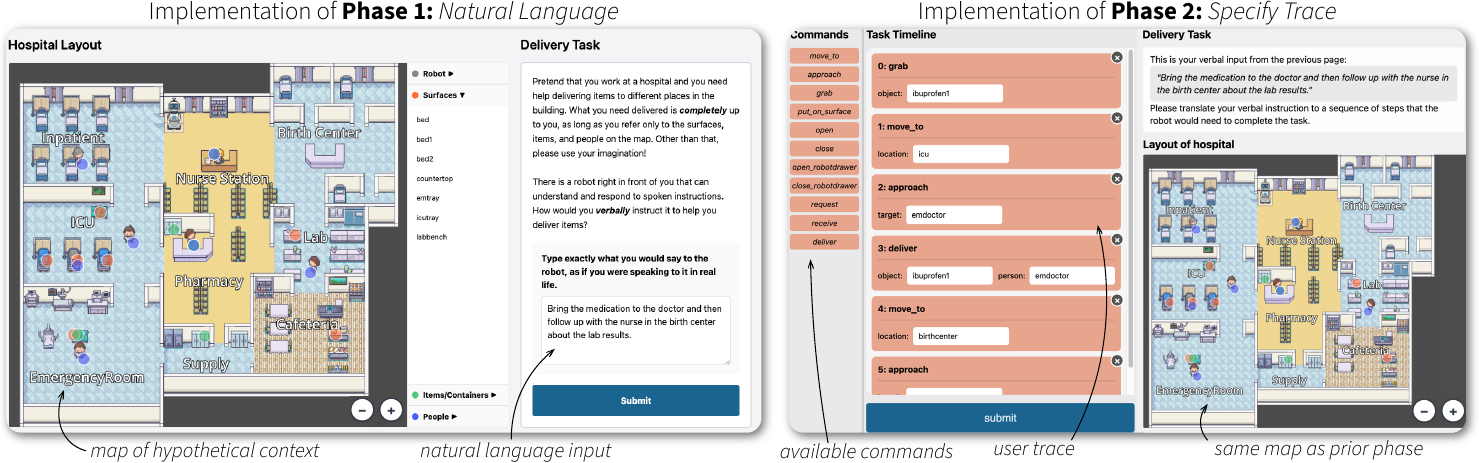}
    \caption{Our implementation of the first and second phases of the \tool{} approach. The map uses graphics from LimeZu \cite{limezu}.}
    \Description{Screenshots of the implemented interface for Phases 1–2, showing fields for entering natural language tasks and resulting task traces, demonstrating how users create, review, and edit early-stage traces.}
    \label{fig:12imp}
\end{figure*}

\subsection{Implementation}\label{sec:implementation}

We implemented the \tool{} approach on a keyboard-and-mouse web-based application consisting of five sequential pages, each corresponding to one \tool{} phase. The entirety of the web interface was constructed via React,\footnote{\href{https://react.dev/}{https://react.dev/}} with its purpose primarily being to serve as the research platform through which we evaluated \tool{} (see Section \S\ref{sec:eval}). We briefly detail the design of each page in order. 

\paragraph{\pone{} Implementation} Figure \ref{fig:12imp} (left) shows our implementation of \pone{} in a web interface. The interface displays a map-based visualization of the robot’s environment, including relevant objects, people, and locations. Users can hover over entities and their locations to inspect them. Users receive a prompt within a pane on the right side of the window and enter a natural language task description below the prompt.

\paragraph{\ptwo{} Implementation} Figure \ref{fig:12imp} (right) shows our implementation of \ptwo{}, where users translate their natural-language instructions into a concrete task trace using a drag-and-drop linear action timeline. Actions can be added, parameterized, reordered, modified, or removed, allowing users to externalize their understanding of how the robot should execute their natural language task description.

\begin{figure}
    \centering
    \includegraphics[width=\columnwidth]{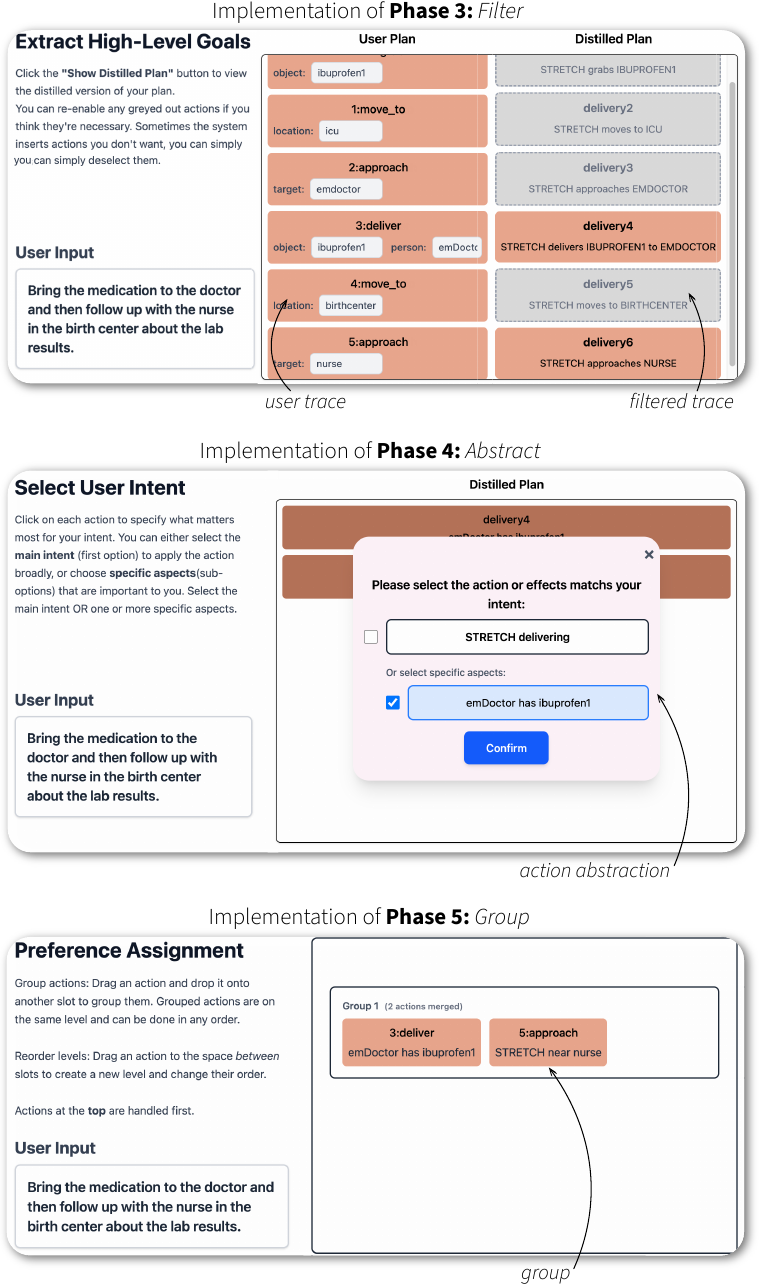}
    \caption{Our implementation of the third, fourth, and fifth phases of the \tool{} approach.}
    \Description{Screenshots of the interface for Phases 3–5, showing tools to mark critical actions, filter steps, and adjust ordering constraints, highlighting interactive refinement of task traces.}
    \label{fig:345imp}
\end{figure}

\paragraph{\pthree{} Implementation} Figure \ref{fig:345imp} (top) illustrates both the user-created trace and the system-filtered trace. System decisions for designating actions as critical and non-critical can be overridden by the user in a step that we term \textit{interactive validation}. The user can simply click on actions to re-classify them depending on what the user believes the robot can infer autonomously.

For making the initial set of filtering decisions, the web interface communicates with a Python webserver backend. The backend accepts the initial trace constructed by the user in \ptwo{} and automatically designates actions as critical or non-critical. While we initially explored using large language models (LLMs) to filter traces, we found that without explicit knowledge of how user intent maps to concrete actions, LLMs often fail to produce reliable, personalized filtered traces. To ensure correctness and eliminate uncertainty in the early testing of our concept, we opted for a classical symbolic approach to the filtering process, which provides the interface with predictable, verifiable results (see Appendix~\ref{sec:appendix-backend} for a description of the algorithm that we used). We plan on exploring LLM-based extensions to \pthree{} in future work.


\paragraph{\pfour{} Implementation} Shown in Figure~\ref{fig:345imp} (middle), \pfour{} enables users to abstract actions to their outcomes---\textit{goals}. In contrast to \pthree{}, this is purely a manual process. For each action in the filtered trace, users click on the action to open a pop-up window displaying its resulting goal predicates. Within the pop-up, users indicate one of two things---whether they prefer the robot to execute the exact action as currently designated, or if achieving the associated outcome alone is sufficient. Thus, users are exposed to the postconditions of critical task outcomes and are allowed to abstract away unnecessary execution details.

\paragraph{\pfive{} Implementation} Shown in Figure \ref{fig:345imp} (bottom), \pfive{} lets users specify execution-order preferences via a drag-and-drop grouping interface. Users can assign actions to priority groups, within which actions and goals are understood to be executed in a nondeterministic fashion. Users can also reorder these groups (if multiple groups exist) to indicate sequencing constraints between groups, with actions in the same group executed in any order or in parallel.


\section{Evaluation}\label{sec:eval}
We conducted an IRB-approved study that assessed the advantages and disadvantages of using \tool{} to guide users to refine their intuitively specified task specifications (natural language and hand-crafted traces) and elicit their ground-truth intent. This study is exploratory in nature with the primary goal \bl{}of answering the research questions outlined in Section \S\ref{sec:intro}, specifically (\textbf{RQ1}) which aspects of the \tool{} approach are effective, or ineffective, at eliciting ground-truth user intent; and (\textbf{RQ2}) how can \tool{} inform the design of future human-robot task communication interfaces?\bk{}

In what follows, we present our study procedure and a description of participants. Then, we present three separate analyses, beginning with the natural language task specifications provided by participants in \pone{}. Following that, we present a mixed-methods analysis of the trace-refinement phases of the \tool{} pipeline, focusing on a quantitative analysis of \ptwo{}, \pthree{}, and \pfour{}, and a qualitative analysis of \pfive{}. Lastly, we provide a qualitative analysis of participant feedback.

\subsection{Study Procedure}
The study was conducted online via Prolific. Each participant completed the study once and responded to a single task prompt using our web interface. We ran two versions of the study---\struct{} and \open{}. We conducted the \struct{} version of the study first, \bl{}the purpose of which was to feed participants a ground-truth task intent and then observe the ability of \tool{} to guide participants toward this ground truth. Specifically,\bk{} participants were given specific task outcomes that the robot needed to achieve in the hospital environment discussed in Section \S\ref{sec:approach}, stated as follows: \qt{The patient must have ibuprofen AND the ICU doctor must have the X-ray file.} We then conducted the \open{} version of the study, in which participants were able to decide their own task outcomes. The procedures for each study were otherwise identical, and environment layouts and item locations were fixed across participants. Each individual participated in only one version of the study. 

After providing informed consent, participants were directed through each phase of \tool{}. Each phase followed the same procedure. First, participants viewed a tutorial video that explained the current phase and how to use the web interface for that phase. The video accompanying the first phase additionally included an introduction to an autonomous robot that is capable of navigation, manipulation, and interaction with humans. Following each video, participants were redirected to the \tool{} web interface. 
%
%
%
At the conclusion of each phase, participants could optionally provide free-form feedback about their experience.

Across all sessions, we collected natural-language instructions, hand-crafted, system-filtered, user-filtered, and abstracted traces, priority groupings, and qualitative feedback. We used all of this information to assess the effectiveness of \tool{}.

\subsection{Participants}
We recruited a total of 61 individuals from Prolific to participate in our study, 30 for the \struct{} condition and 31 for the \open{} condition. Our recruitment criteria were such that participants were (a) between the ages of 18 and 100; (b) located geographically in the United States; (c) not using mobile devices to participate; and (d) had a Prolific approval rating between 98 and 100. Three participants provided only natural language data due to leaving the study early (one in each condition) or providing nonsensical data in later phases of the study (one \struct{}). These participants were compensated for their time, but only provided partial data.

\subsection{Analysis of Natural Language}
In addressing our research questions, we began with an analysis of the natural language that participants provided in \pone{}. Specifically, we were interested in understanding whether and how token length and the lexical features of participant input support \tool{}'s design choices in our pipeline (\textbf{RQ1}) and the broader implications that these lexical features have on human-robot communication tools (\textbf{RQ2}). Lexical features of interest include Sequence terms (\eg{} \qt{deliver medication \textbf{then} linens}) and Conditionals (\eg{} \textit{approach the doctor \textbf{if}~she is there}). A third lexical feature, Step, refers to the presence of multiple independent commands in a user's natural language input, marked by punctuation and conjunctions (\eg{} \qt{deliver the medication \textbf{and grab} the linens}). 

\subsubsection{Measures and Analysis}
Our measures include token length and the frequency of lexical features that appear between the \struct{} and \open{} studies. Crucially, we were able to detect each of these features through regular expressions (see Appendix ~\ref{sec:appendix-reg} for the regular expressions that we used). Readers should note that these results should be interpreted as \textit{associational} rather than \textit{causal} due to the sequential nature in which each study was run (\ie{} participants were not randomly assigned to different studies). We then use independent t-tests to compare mean token length and $\chi^2$ tests to compare the frequency of lexical features.

\begin{figure*}
    \centering
    \includegraphics[width=\textwidth]{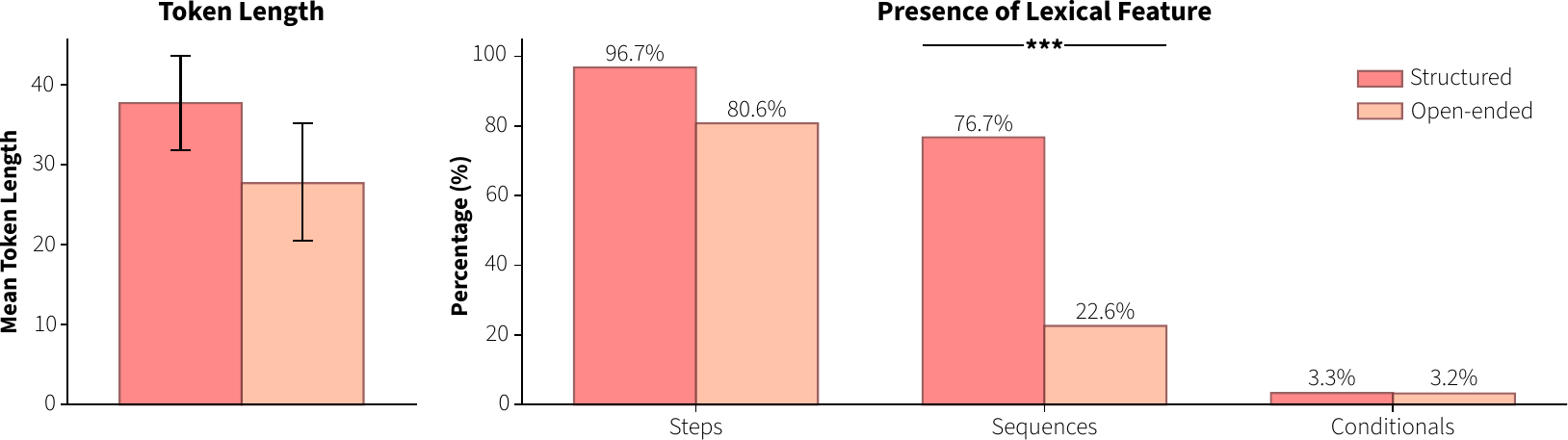}
    \caption{Comparison of natural language input length (left) and of the different lexical features occurring (right) between the \struct{} and \open{} studies in \pone{}. Error bars are standard error. †p<0.1, *p<0.05, **p<0.01, ***p<0.001}
    \Description{Two-panel plot comparing Phase 1 natural language input between structured and open-ended studies. The left panel shows input lengths; the right panel shows the frequency of three lexical features, with significance markers indicating differences, including higher sequence markers in structured tasks.}
    \label{fig:nl_eval}
\end{figure*}

\subsubsection{Results} Figure \ref{fig:nl_eval} depicts the results of our natural language evaluation. Participants in the \struct{} and \open{} studies provided natural language input with a mean length of $38.23$ (SD=$33.45$) and $28.10$ (SD=$41.28$) tokens, respectively. This difference was not significant. Participants in the \struct{} study utilized the Steps, Sequences, and Conditionals lexical features $96.7\%$, $76.7\%$, and $3.3\%$ of the time, respectively, whereas participants in the \open{} study utilized these same features $80.6\%$, $22.6\%$, and $3.2\%$ of the time, respectively. We only detected a significant difference in the frequency of Sequences occurring, $\chi^2(1)=15.7$, $p<0.001$. 

These results indicate that sequence markers are the most discriminative linguistic feature distinguishing \struct{} from \open{} instructions. \textit{Structured} instructions are significantly more likely to include explicit ordering cues such as “then,” “next,” and “after.” In contrast, realistic uses of the \tool{} approach are likely to resemble \open{} instructions, which often omit explicit sequencing. This absence of sequence markers makes it ambiguous whether participants intend a specific ordering of actions, thereby directly addressing \textbf{RQ2} by underscoring the need for human-robot communication interfaces to adopt processing pipelines that can infer or elicit users’ underlying intent.

\subsection{Analysis of Trace Refinement (\ptwo{} through \pfive{})}

In order to further answer \textbf{RQ1}, we conducted a multifaceted quantitative analysis of \ptwo{}, \pthree{}, and \pfour{}, and a qualitative analysis of \pfive{}. For the third phase, we additionally distinguished between filtered traces that were produced by the system before being modified by the user (\textit{system-filtered}) versus these same traces after being modified by the user (\textit{user-filtered}). We analyzed traces from 28 \struct{} (two discarded due to missing or nonsensical data) and 30 \open{} (1 discarded due to missing data) participants. Additional exclusions occurred for specific analyses, which we note where applicable.

\subsubsection{Measures and Analysis}

We list our measures below:
\begin{itemize}
    \item \textit{Goal achievement}---this measure applies only to the \struct{} study and measures whether the participant-created traces successfully resulted in the \struct{} objectives, namely that the patient received ibuprofen and the doctor received the X-ray files.
    \item \textit{Trace length}---this measure applies to both studies and measures the length of participant traces at the end of \ptwo{}, \pthree{}, and \pfour{}. Shorter traces over the course of progressing through each phase indicate that \tool{} successfully encourages participants to remove steps from their traces. Note that \pfour{} does not allow users to change the length of their traces; thus, the trace length at the end of this phase is the same as the trace length at the end of \pthree{} (the \textit{user-filtered} traces). Note that for the \struct{} study, we only analyze traces that achieve the study objectives (n=21).
    \item \textit{Plan length}---this measure applies to both studies and measures the length of \textit{task plans} produced from participant traces in \ptwo{}, \pthree{}, and \pfour{}. Recall that a \textit{plan} refers to the steps that a robot will perform in order to execute a hand-crafted trace. For example, if the user's trace is (1) \cmd{deliver}{ibuprofen, patient} followed by (2) \cmd{approach}{nurse}, a resulting plan might involve the robot (1) moving to the ibuprofen, (2) grabbing it, (3) moving to the patient, (4) giving it to the patient, and then (5) approaching the nurse. Plans are computed via the Interaction Specification Language \cite{porfirio2025interaction}. Longer plans are indicative of mistakes made during trace creation and refinement, leading to backtracking actions. The \struct{} study only includes plans that achieve study objectives (n=21). For the \open{} study, we exclude one participant due to failed plan computation in \pfour{} (n=29).
    \item \textit{Plan length in novel environments}---this measure applies to both studies with the aim of evaluating trace robustness under environmental uncertainty. To compute this measure, we randomize initial item placements while keeping task goals fixed (discarding placements that prevent a plan from being computed) and calculate task plans. Longer perturbed plans are indicative of the robot being over-constrained. The \struct{} study similarly only includes plans that achieve study objectives (n=21), and the \open{} study similarly excludes the participant whose plan failed to compute (n=29).
\end{itemize}

The grouping patterns in \pfive{} are measured qualitatively due to the difficulties in applying the above measures to the inherently nondeterministic group structures produced by participants. In particular, we examine language-structure alignment by analyzing how linguistic cues in the natural-language instructions, including sequential and grouping keywords, correspond to users’ specified priority group structures.

For our analysis of trace length, plan length, and perturbed plan lengths, we use Friedman omnibus tests with \textit{post hoc} Wilcoxon tests with Bonferroni correction to assess pairwise differences. In applying the Bonferroni correction, we multiply our p-values by the number of comparisons. 
We employed non-parametric statistical methods due to the high prevalence of outliers in our data. 

\subsubsection{Goal Achievement Results}
In this section, we evaluate only the \struct{} study condition, as ground-truth goals are not available for the \open{} study. Each \struct{} study required participants to achieve two explicit outcomes. We compare user-selected intent predicates with actual planner outcomes to assess goal articulation accuracy. 

System-filtered traces achieved full goal completion in 21 of 28 cases (75.0\%), partial completion in 5 of 28 cases (17.9\%), and no goal completion in 2 of 28 cases (7.1\%). In total, distilled traces achieved 47 of 56 goals, corresponding to an overall goal achievement rate of 83.9\%. User-filtered traces and abstracted traces exhibited the same distribution of full, partial, and non-achievement, indicating that automated distillation was sufficient to preserve goal satisfaction in the \struct{} study. Subsequent user validation primarily served to confirm intent interpretation and refine action abstraction rather than improving overall goal completion. 

\subsubsection{Trace Length Results}
\begin{figure*}
    \centering
    \includegraphics[width=\textwidth]{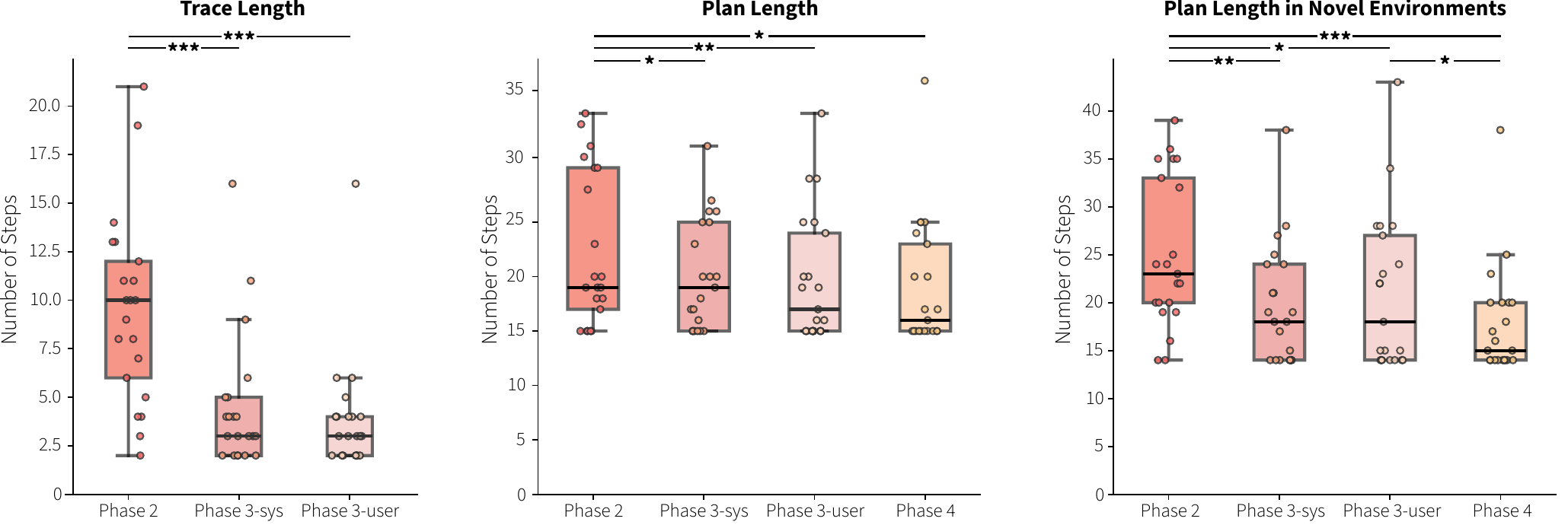}
    \caption{Trace and plan length for the \struct{} study (n=21). Lower values are better. †p<0.1, *p<0.05, **p<0.01, ***p<0.001}
    \Description{Box plots of trace and planner output lengths for four representation types in the structured study: user-created, system-filtered, user-filtered, and abstracted. Medians, variability, and significant differences are indicated with †, *, **, *** markers.}
    \label{fig:tracelength_st}
\end{figure*}

We compare the number of actions per trace across phases two through four. Figures~\ref{fig:tracelength_st} and~\ref{fig:tracelength_oe} (left panel) depict trace length results for \struct{} and \open{} tasks, respectively. Table~\ref{tab:stats-complete} provides detailed statistical comparisons.

\textbf{Structured study.} User-created traces contained $M = 9.52$ actions ($SD = 4.91$), while system-filtered traces reduced this to $M = 4.52$ actions ($SD = 3.52$), and user-filtered traces contained $M = 3.90$ actions ($SD = 3.05$). System distillation achieved a 52.5\% reduction, with a net reduction of 59.0\% after user validation. Both reductions were highly significant ($p < 0.001$), with no significant difference between system-filtered and user-filtered traces.

\textbf{Open-ended study.} User-created traces contained $M = 5.93$ actions ($SD = 4.13$), while system-filtered traces reduced this to $M = 3.10$ actions ($SD = 2.90$), and user-filtered traces contained $M = 3.23$ actions ($SD = 3.34$). System distillation achieved a 47.7\% reduction, with a net reduction of 45.5\% after user validation. Both reductions were highly significant ($p < 0.001$), with no significant difference between system-filtered and user-filtered traces.

\bl{}\textbf{Practical significance: } Across both task conditions, \tool{} cuts trace length by approximately half\bk{}. This comparison captures how much redundancy is removed by the system and how little is reintroduced by users during interactive validation. Minimal user modifications suggest high agreement with \tool{}.

\small
\begin{table}[t]
\centering
\small
\Description{Statistical comparisons of trace and plan lengths for structured and open-ended studies. System-filtered and user-filtered traces generally have reduced lengths compared to user-created traces.}
\begin{tabular}{llccl}
\toprule
\textbf{Study} & \textbf{Comparison} & \textbf{Statistic} & \textbf{p-value} & \textbf{Effect} \\
\midrule

\multicolumn{5}{l}{\cellcolor[HTML]{D5D8DC}\textcolor{black}{\textbf{\textit{Measure: Trace Length}}}} \\

Structured & Omnibus  & $\chi^2(2) = 34.94$ & $< 0.001$ & \\
& Ph2 vs Ph3-\textsf{\textbf{s}}  & $W = 0.0$ & $< 0.001$ & $\vee$52.5\%  \\
& Ph2 vs Ph3-\textsf{\textbf{u}}  & $W = 0.0$ & $< 0.001$ & $\vee$59.0\%  \\
& Ph3-\textsf{\textbf{s}} vs Ph3-\textsf{\textbf{u}}  & $W = 6.0$ & 0.605 & \\
\cmidrule{2-5}
Open-ended & Omnibus  & $\chi^2(2) = 43.63$ & $< 0.001$ & \\
& Ph2 vs Ph3-\textsf{\textbf{s}}  & $W = 0.0$ & $< 0.001$ & $\vee$47.7\%  \\
& Ph2 vs Ph3-\textsf{\textbf{u}}  & $W = 0.0$ & $< 0.001$ & $\vee$45.5\%  \\
& Ph3-\textsf{\textbf{s}} vs Ph3-\textsf{\textbf{u}}  & $W = 49.0$ & 1.000 & \\
\midrule

\multicolumn{5}{l}{\cellcolor[HTML]{D5D8DC}\textcolor{black}{\textbf{\textit{Measure: Plan Length (Original Environment)}}}}\\

Structured & Omnibus  & $\chi^2(3) = 22.94$ & $< 0.001$ & \\
& Ph2 vs Ph3-\textsf{\textbf{s}}  & $W = 0.0$ & 0.022 & $\vee$10.0\%  \\
& Ph2 vs Ph3-\textsf{\textbf{u}}  & $W = 2.0$ & 0.010 & $\vee$11.6\%  \\
& Ph2 vs Ph4  & $W = 8.0$ & 0.020 & $\vee$14.5\%  \\
& Ph3-\textsf{\textbf{s}} vs Ph3-\textsf{\textbf{u}}  & $W = 21.0$ & 1.000 & \\
& Ph3-\textsf{\textbf{s}} vs Ph4  & $W = 16.0$ & 0.456 & \\
& Ph3-\textsf{\textbf{u}} vs Ph4  & $W = 4.0$ & 0.612 & \\
\cmidrule{2-5}
Open-ended & Omnibus  & $\chi^2(3) = 14.05$ & 0.003 & \\
& Ph2 vs Ph3-\textsf{\textbf{s}}  & $W = 0.0$ & 1.000 & \\
& Ph2 vs Ph3-\textsf{\textbf{u}}  & $W = 5.5$ & 0.094 & \\
& Ph2 vs Ph4  & $W = 21.0$ & 0.304 & \\
& Ph3-\textsf{\textbf{s}} vs Ph3-\textsf{\textbf{u}}  & $W = 5.5$ & 0.287 & \\
& Ph3-\textsf{\textbf{s}} vs Ph4  & $W = 20.5$ & 0.513 & \\
& Ph3-\textsf{\textbf{u}} vs Ph4  & $W = 5.0$ & 1.000 & \\
\midrule
\multicolumn{5}{l}{\cellcolor[HTML]{D5D8DC}\textcolor{black}{\textbf{\textit{Measure: Plan Length (Perturbed Environment)}}}} \\

Structured & Omnibus  & $\chi^2(3) = 33.90$ & $< 0.001$ & \\
& Ph2 vs Ph3-\textsf{\textbf{s}}  & $W = 1.5$ & 0.002 & $\vee$21.8\%  \\
& Ph2 vs Ph3-\textsf{\textbf{u}}  & $W = 18.0$ & 0.021 & $\vee$16.5\%  \\
& Ph2 vs Ph4  & $W = 0.0$ & $< 0.001$ & $\vee$29.2\%  \\
& Ph3-\textsf{\textbf{s}} vs Ph3-\textsf{\textbf{u}}  & $W = 22.0$ & 0.642 & \\
& Ph3-\textsf{\textbf{s}} vs Ph4  & $W = 9.0$ & 0.118 & \\
& Ph3-\textsf{\textbf{u}} vs Ph4  & $W = 4.0$ & 0.039 & $\vee$15.2\%  \\
\cmidrule{2-5}
Open-ended & Omnibus  & $\chi^2(3) = 24.77$ & $< 0.001$ & \\
& Ph2 vs Ph3-\textsf{\textbf{s}}  & $W = 0.0$ & 0.009 & $\vee$9.6\%  \\
& Ph2 vs Ph3-\textsf{\textbf{u}}  & $W = 19.5$ & 0.014 & $\vee$14.7\%  \\
& Ph2 vs Ph4  & $W = 11.5$ & 0.013 & $\vee$17.4\%  \\
& Ph3-\textsf{\textbf{s}} vs Ph3-\textsf{\textbf{u}}  & $W = 37.0$ & 1.000 & \\
& Ph3-\textsf{\textbf{s}} vs Ph4  & $W = 20.5$ & 0.513 & \\
& Ph3-\textsf{\textbf{u}} vs Ph4  & $W = 23.0$ & 1.000 & \\
\bottomrule
\end{tabular}
\caption{Complete statistical analysis results for trace and plan length comparisons across \struct{} and \open{} studies. All Wilcoxon p-values are Bonferroni-corrected. Phase is designated by \textit{Ph2}, \textit{Ph3}, and \textit{Ph4}, with \textit{Ph3-s} being system-filtered traces and \textit{Ph3-u} being user-filtered traces.}
\label{tab:stats-complete}
\end{table}
\normalsize

\subsubsection{Plan Length Results}
Using a task planner, we compute a plan from each trace and measure the number of steps required. We compare this measure across phases to assess whether distilled traces lead to shorter (more efficient) plans. Figures~\ref{fig:tracelength_st} and~\ref{fig:tracelength_oe} (middle) depict plan length in the original environment. Detailed statistical comparisons are provided in Table~\ref{tab:stats-complete}.

\textbf{Structured study.} Plans generated from user-created traces required $M = 22.29$ steps ($SD = 7.05$), compared to $M = 20.05$ steps ($SD = 5.06$) for system-filtered traces, $M = 19.71$ steps ($SD = 5.89$) for user-filtered traces, and $M = 19.05$ steps ($SD = 5.87$) for abstracted traces. User-created traces required significantly more steps than all other traces ($p < 0.05$ for all comparisons, with reductions ranging from 10.0\% to 14.5\%). Differences among system-filtered, user-filtered, and abstracted traces were not significant, suggesting these traces converge on similar levels of efficiency.

\textbf{Open-ended study.} Plans from user-created traces required $M = 10.69$ steps ($SD = 7.73$), compared to $M = 10.41$ steps ($SD = 7.35$) for system-filtered traces, $M = 9.83$ steps ($SD = 7.49$) for user-filtered traces, and $M = 9.59$ steps ($SD = 7.63$) for abstracted traces. While the overall Friedman test revealed significant differences across phases ($\chi^2(3) = 14.05$, $p = 0.003$), pairwise comparisons with Bonferroni correction did not reach significance. The smaller baseline plan lengths in the \open{} study (M = 10.69 steps) may limit the magnitude of detectable pairwise improvements, though the overall pattern of optimization across phases remains evident.

\bl{}\textbf{Practical significance: } In the \struct{} study, \tool{} yields efficiency gains early in the pipeline, with most improvements emerging by \pthree{} and no significant gains in subsequent phases. This pattern highlights filtration as the primary driver of efficiency within \tool{}. However, these gains are modest (10–15\%) and do not replicate in the \open{} study, suggesting that \tool{}’s ability to improve plan efficiency may depend on the presence of well-defined or externally specified objectives.\bk{}

\subsubsection{Plan Length for Novel Environments Results}
\begin{figure*}
    \centering
    \includegraphics[width=\textwidth]{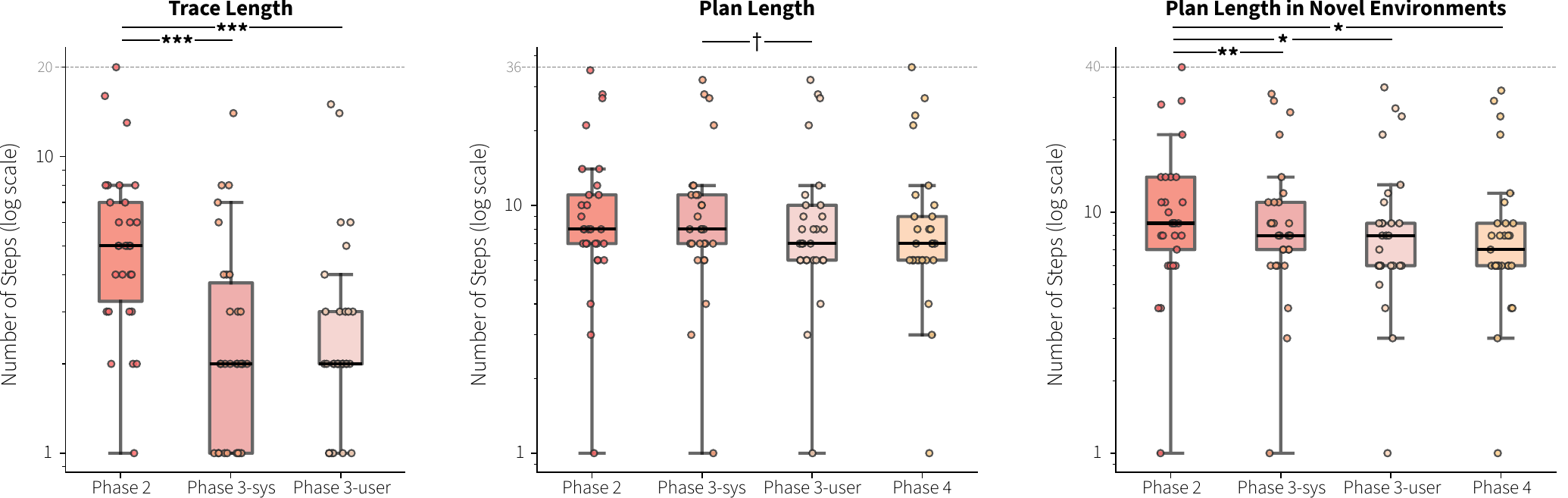}
    \caption{Trace and plan length on a logarithmic scale for the \open{} study, where n=30 for trace length comparisons and n=29 for plan length comparisons.  Lower values are better. This study exhibited higher variance (as evidenced by the prevalence of outliers) than the \struct{} study. †p<0.1, *p<0.05, **p<0.01, ***p<0.001}
    \Description{Box plots of trace and planner output lengths on a logarithmic scale for four representation types in the open-ended study: user-created, system-filtered, user-filtered, and abstracted. Medians, variability, and significant differences are indicated with †, *, **, *** markers.}
    \label{fig:tracelength_oe}
\end{figure*}

To evaluate the performance of traces in novel environments, we randomized item placement and initial conditions in the environment while keeping task goals fixed. For each randomized configuration, we execute plans from all four traces. Figures~\ref{fig:tracelength_st} and~\ref{fig:tracelength_oe} depict plan length in perturbed (right) versus original environments (middle). All statistical test results are reported in Table~\ref{tab:stats-complete}.

\textbf{Structured study.} In novel environments, user-created traces required $M = 25.10$ steps ($SD = 7.82$), while system-filtered traces required $M = 19.62$ steps ($SD = 6.28$), user-filtered traces required $M = 20.95$ steps ($SD = 8.10$), and abstracted traces required $M = 17.76$ steps ($SD = 5.73$). User-created traces required significantly more steps than all other traces, with reductions ranging from 16.5\% to 29.2\% (all $p < 0.05$).

A purely descriptive comparison between original and perturbed plan lengths indicates an increase of 12.6\% for user-generated traces (an average length of 22.29 in the original environment versus an average length of 25.10 in the perturbed environment), showing brittleness to environmental changes. In contrast, system-filtered plan length improves by 2.1\% (20.05 $\rightarrow$ 19.62 steps), user-filtered plan length increases by only 6.3\% (19.71 $\rightarrow$ 20.95 steps), and abstracted plan length improves by 6.8\% (19.05 $\rightarrow$ 17.76 steps).


\textbf{Open-ended study.} In perturbed environments, user-created traces required $M = 11.52$ steps ($SD = 8.35$), compared to $M = 10.41$ steps ($SD = 7.29$) for system-filtered traces, $M = 9.83$ steps ($SD = 7.41$) for user-filtered traces, and $M = 9.52$ steps ($SD = 7.52$) for abstracted traces. User-created traces required significantly more steps than all other traces, with reductions ranging from 9.6\% to 17.4\% (all $p < 0.05$). User-created plans increased by 7.8\% from original to perturbed environments (10.69 $\rightarrow$ 11.52 steps), while system-filtered and user-filtered plans remained stable with no change (both 10.41 and 9.83 steps in both environments), and abstracted plans slightly improved by 0.7\% (9.59 $\rightarrow$ 9.52 steps). 

\bl{}\textbf{Practical significance: }
Across both studies, \tool{} produces traces that generalize more robustly to novel environmental configurations, with distilled and abstracted traces both maintaining and improving performance under variation. While \pfour{} yields a modest improvement over \pthree{} in the \struct{} condition, this effect does not consistently appear across other comparisons. Taken together, these findings suggest that the primary gains in robustness stem from filtration, with abstraction providing only limited additional benefit.\bk{}

\subsubsection{Ordering Preferences and Task Grouping Results}


We systematically analyzed the correspondence between linguistic cues in users' natural language inputs and their actual priority group structures. Linguistic cues were categorized into two types: (1) \textit{sequential} keywords indicating ordered execution (e.g., ``first,'' ``then,'' ``after''), and (2) \textit{grouping} keywords suggesting parallel or flexible ordering (e.g., ``and,'' ``both,'' ``together'').\bl{} See Appendix~\ref{sec:appendix-reg} for the full list of lexical cues and regex patterns.\bk{}

We computed alignment by comparing expected priority structures based on linguistic cues against actual user-defined groups, scored across two dimensions. \textbf{Grouping alignment}: when users employed grouping keywords (e.g., ``deliver X \textbf{and} deliver Y''), we expected at least one instance of grouped actions to appear in \pfive{} to allow for execution in a nondeterministic order. \textbf{Sequential alignment}: when users employed sequential keywords (e.g., ``deliver X, \textbf{then} deliver Y''), we expected there to exist at least one instance of separate priority groups with deterministic ordering. Alignment was measured at the participant level; a participant was counted as \textit{aligned} if a keyword and its corresponding priority structure co-occurred anywhere in the participant’s response, regardless of where that structure appeared in the overall grouping.

\textbf{Structured study.} We identified a total of 33 sequential keywords and 53 grouping keywords across all natural language commands provided by participants. The alignment rate, measuring the existence of keywords that matched any priority group, reached 85.7\%, with 24 of 28 participants achieving at least one cue-structure match. The 53 grouping keywords outnumbered the 33 sequential keywords, reflecting an interesting tendency towards organizing related information into batches. The frequency of these cue types, combined with the high alignment rate, demonstrates that users employed both grouping and sequencing strategies thoughtfully and consistently.

\textbf{Open-ended study.} In contrast, this study showed lower alignment. Among the 30 natural language commands paired with priority group structures, 22 contained at least one sequential or grouping cue (11 sequential and 41 grouping cues overall). Only 13 of 30 (43.3\%) showed at least one cue-structure match. 
Although many open-ended participants used grouping or sequential language, fewer translated those into explicit priority organization. This suggests that structured task framing better supported participants in expressing ordering intent through formal priority groups, whereas in open-ended commands, linguistic cues were more often used descriptively rather than as precise indicators of execution structure.


\subsubsection{Additional Analysis: User vs. System Judgments of Essentiality}

We compare system-filtered plans with user-filtered distilled plans to understand how users reinterpret action essentiality. Table~\ref{tab:user-system-alignment} summarizes how participants responded to system filtration decisions, reporting the proportion of participants who made no changes, reselected at least one previously filtered action, or deselected at least one system-retained action. In the \struct{} study, users agreed with system judgments 57.14\% of the time, reselected previously removed actions 17.86\% of the time, and deselected system-retained actions 35.71\% of the time. In the \open{} study, agreement was lower at 43.33\%, with higher reselection rates (40.0\%) and deselection rates (30.0\%). The \struct{} task's higher agreement rate suggests that predefined action sets facilitate clearer communication of task requirements, while the \open{} study's higher reselection rate indicates users felt the system over-pruned demonstrations. These differences reveal context-dependent mismatches between algorithmic distillation and human judgment. 

\begin{table}[t]
\centering
\small
\Description{Alignment between system-filtered and user-filtered judgments of action essentiality. Structured tasks: 65.5\% agreement, 6.9\% reselection, 27.6\% deselection. Open-ended tasks: 53.3\% agreement, 26.7\% reselection, 20.0\% deselection.}
\begin{tabular}{lccc}
\toprule
\textbf{Task Type} & \textbf{No Change} & \textbf{Reselect Any} & \textbf{Deselect Any} \\
\midrule
\struct{} study & 57.14\% & 17.86\% & 35.71\%   \\
\open{} study  & 43.33\% & 40.0\% & 30.0\% \\
\bottomrule
\end{tabular}
\caption{Alignment between system-filtered and user-filtered judgments of action essentiality across task types.}
\label{tab:user-system-alignment}
\end{table}

\subsection{Participant Feedback}\label{sec:feedback}
To help further answer both \textbf{RQ1} and \textbf{RQ2}, we additionally conducted a qualitative analysis of user experience based on the feedback offered by participants at the end of each stage of the \tool{} pipeline. This analysis followed a Reflexive Thematic Analysis (RTA) approach \cite{braun2021thematic}, in which the last author (a) familiarized themselves with the data; (b) developed an initial set of codes and coded each instance of feedback provided by participants; and then (c) organized codes into themes that emerged from this process. 

A total of 24 participants provided qualitative feedback. Feedback occurred at any stage of \tool{}; some participants provided feedback at all stages, while others provided feedback at only a single stage. In our thematic analysis of this feedback, 16 codes emerged, which we grouped into three themes. We report on each of those themes below, referring to participants using the ``PXY'' identifier, where X is the participant ID that ranges from 1-61 and Y is whether or not the participant engaged in the \struct{} or \open{} study. 

\subsubsection{Theme: Robot Capabilities}

This theme pertains to how views of the robot's capabilities and autonomy affected user input to \tool{}. In the first and second phases, P28S and P31O expressed that the robot's autonomy affected their initial input, with P28S saying that the task \qt{was easier, understanding that the robot knew how to do some things automatically} and P31O saying that \qt{since it is autonomous, I am assuming it can find its own way.} One other participant, however, indicated low trust in the robot's autonomy, which they attributed to precisely restricting the robot to an exact sequence of actions---\qt{I am very specific because if I'm not, people often mess up, and I imagine a robot will mess up even more}, and thus, \qt{the procedure is more important than the end result for me} (P7S). This participant was therefore hesitant to relinquish control to the robot in deciding which actions it needed to perform.

Three other individuals expressed uncertainty in the robot's autonomy (P8S and P38O) and its capabilities and limitations (P8S, P25S). P8S stated, \qt{I ... wondered how much autonomy it had}, while P25S stated, \qt{Not sure what the robot can understand or if it knows where different rooms are, or even if it knows who the dr is.}

\paragraph{Implications.}  Different interpretations of the robot's autonomy, capabilities, and limitations may affect individual outcomes from the \tool{} approach. Even individuals who interpret the robot's autonomy similarly vary in their willingness to relinquish control to the robot. \citet{zhang2025balancing} show that end-user programming interfaces affect user perceptions of robot autonomy, but our result shows that more work is needed in uncovering how perceptions of autonomy affect interface usage, and thus, how interfaces should be designed to accommodate different perceptions of autonomy. 

\subsubsection{Theme: Preference}\label{sec:preferencethemee}

A key goal of the \tool{} approach is eliciting users' ground truth task intent, in which ground truth refers to a user's true preference of actions and objectives, and their order. Even in the \struct{} study, some participants indicated that our ``ground truth'' was affected by their preferences for how they believed the robot should perform the task, despite this ground truth being explicitly fed to them as an unordered set of task objectives. Shedding light on their interpretation, P23S stated, \qt{I already ordered the tasks specifically from the get-go. I think it's more essential that a patient gets their medication first, and then objects like files and equipment can go to the staff afterwards.} Although P7S did not enforce the same ordering constraint, they admitted that a different layout of the hospital environment may have led them to enforce such an order: \qt{The nurse station and the pharmacy are basically in the same general area, so it doesn't matter to me whether the robot grabs the [X-ray] or the medicine first.}

In the \open{} study, two participants (P37O and P38O) similarly expressed a substantial preference for how the robot should perform its task. For example, P38O stated a strong preference for medication being delivered to the doctor rather than a patient: \qt{In order to avoid liability, it is important for the [medication] to be delivered directly to the doctor unless there are other options.} P37O, by contrast, indicated a softer preference, with some room for interpretation by the robot: \qt{They only need to approach the Nurse. That's the most important step. Anything else isn't necessary.}

\paragraph{Implications.} Even if task objectives are fed directly to users for conveying to the robot, users demonstrate a tendency to enforce their preferences on these objectives. It is therefore important for task specification interfaces to be designed with the ability for users to express preferences. This aligns with existing guidelines for human-AI interaction design \cite{amershi2019guidelines}. Based on our work, we highlight the need for future work in extending this design principle to natural language interfaces, which disproportionately place the burden of interpreting ambiguous user intent on the robot, while offering little support for the user to clarify their preferences.

\subsubsection{Theme: User Experience}

Participants expressed having both positive and negative experiences with the \tool{} approach and with the web interface. Common feedback included that the interface and prompt were \qt{easy} (P7S, P38O, P45O, P46O, P52O). In \ptwo{}, P3S indicated that hand-crafting a trace made it \qt{simpler to think about the natural language given the categories available}. Several other participants (P3S, P4S, P9S, P11S, P51O, P54O, P57O) indicated that they had a good experience overall.

Some participants did not have as positive an experience, with P8S, P10S, P23S, and P47O highlighting usability concerns in \ptwo{}. One such concern related to the ease of dragging actions into the Task Timeline in this phase, with P10S saying, \qt{Once the number of steps reaches the height of that UI element, adding new steps to the bottom is more difficult than it should be.} P8S echoed this statement. Another concern pertained to backtracking to prior phases (P23S and P47O), with P23S indicating that \qt{the study advanced too early after clicking submit} and that they \qt{wanted to alter the steps that I inputted.} While backtracking is possible within our web interface, it clearly was not a seamless experience for some participants.

Other participants had a negative experience because it was not possible to express their desired task in terms of the actions available to them (P45O), or because the task was \qt{difficult} (P52O) or \qt{tricky} (P56O). One participant reported a potential bug in the interface (P61O), specifically that they could not select or deselect actions in \pthree{}. We were not able to recreate this on our own devices. Because this participant was still able to complete the study, we retained their data.

Lastly, several participants indicated a desire for additional functionality in our web interface. P38O, in particular, wanted the interface to support branching, rather than purely linear traces (\qt{there should be alternatives of what to do in certain areas in the command}). Other participants (P38O and P45O) expressed a desire to see how the robot would respond to a natural language request, \ie{} via simulation, with P38O saying, \qt{I would have liked to have practiced ... to see what the robot would actually do.} 

\paragraph{Implications.} We find it encouraging that many participants expressed having a positive experience with \tool{}. Our operationalization of \tool{} on the web interface received a mix of feedback, which helps inform design recommendations for future tools that implement \tool{}. Most notably, these tools should facilitate easy navigation between phases. Supporting branching is additionally a natural extension to the \tool{} approach, and would enable \tool{} to mirror the expressiveness of standard end-user programming tools. Lastly, interfaces that adopt the \tool{} approach may consider whether integrating a robot simulator is appropriate. If \tool{} is meant to be used in scenarios requiring rapid task specification, viewing the robot on a simulator in the \pone{} or \ptwo{} phases may be too time-consuming and distract users from the benefits of later \tool{} phases. Indeed, the purpose of \tool{} is to encourage ground truth preference elicitation \textit{irrespective} of any robot or simulator. Still, a simulator may present certain benefits, such as by aligning user expectations of robot performance with actual robot task execution. In this way, a simulator may serve as motivation to complete the later steps of \tool{} in order to further this alignment.



\section{Discussion}
In answering our research questions, we reflect on \bl{}the practical findings that emerged from our results\bk{} and discuss design implications for future human-robot task communication interfaces informed by both quantitative and qualitative results. Then, we acknowledge the limitations of our work while offering future research directions.

\subsection{From Traces to Intent}
Our results show that raw traces alone (\ptwo{}) often include significantly more steps than the filtered and abstracted traces in \pthree{} and \pfour{}. While in most cases (75\% of the time in the \struct{} study) these traces ultimately lead the robot to correctly achieve its goal, \bl{}our results on resulting plan length show that the existence of\bk{} extra steps adds unnecessary detail that may lead the robot to perform its task significantly less efficiently, especially when the locations of entities in the environment change. \bl{}This is consistent with prior work showing that users intuitively insert non-critical actions into their task specifications \cite{porfirio2025interaction, porfirio2023crowdsourcing, porfirio2024goal}, and that the essential steps of a task are often hidden within a longer sequence of environment-dependent instructions \cite{liao2019synthesizing}. Actions preserved through distillation aligned with information necessary to capture the user's task constraints and preferences, whereas removed actions reflected details beyond what was required for a sufficient and minimal expression of intent\bk{}. \tool{} \bl{}achieves this\bk{} by distilling user-created traces into compact representations of critical actions.

\bl{}The \struct{} study shows that in addition to being more efficient, distilled traces more closely matched the dual-outcome ground-truth intent that was fed to participants. This is particularly apparent in our measure of trace length, where participants' distilled traces were significantly more closely aligned in length to the dual-outcome intent than their original traces. We therefore have evidence that \tool{} guides users closer to a ground truth. As evidenced by Section \ref{sec:preferencethemee} (\textit{Theme: Preference}), however, participants' own ground truth may differ from that which we gave them. Thus, future work is needed to see the degree to which \tool{} can help participants achieve their own preferences.\bk{}


\bl{}
\textit{\textbf{Implication: future EUP tools can benefit from \tool{}.}} The observed progression from user-created to distilled traces suggests that \bl{}the \tool{} approach may be effective for users of EUP systems, who are often faced with enumerating or demonstrating the explicit sequence of step-by-step actions for the robot to follow. Through \tool{}'s\bk{} iterative, collaborative process with the user in the same vein as mixed-initiative interfaces \cite{horvitz1999principles},\bl{} users of EUP tools can create more robust traces that more effectively align with their intent.\bk{}

\subsection{From Natural Language to Intent}
\bl{}We observed users variably expressing preferences about how, when, and in what order tasks should be executed. Preferences are more explicitly communicated in the \struct{} study, where we observed a substantially greater presence of lexical features that communicate explicit sequentiality (76.7\%) compared to the \open{} study (22.6\%). The \struct{} results are therefore consistent with prior work that characterizes human-robot communication as explicit, direct, and incremental \cite{marge2020let}. We also observed greater alignment between linguistic cues and priority groups in the \struct{} study (85.7\%) relative to the \open{} study (43.3\%). Though future work is necessary to validate this finding, it is possible that the \struct{} study scaffolded users toward trace-level articulation, encouraging them to externalize ordering constraints. In contrast, the \open{} study appears to elicit more underspecified ordering constraints. Overall, the ambiguity present in the \open{} study is consistent with prior work \cite{piantadosi2012communicative, briggs2017enabling, zhang2025can}, though it is unclear why differences emerged between the two studies. Future work is necessary to further characterize these differences.\bk{}

\textit{\textbf{Implication: the \tool{} approach is well-aligned with the structure of sequential natural language instructions.}} As evidenced in our results, users sometimes provide sequential instructions to the robot. If each instruction is translated to an individual command, this will result in a trace similar to that provided by the user in \ptwo{}, from which the \tool{} approach can assist in removing non-critical actions. It is unclear, however, how well the \tool{} approach generalizes to settings where instructions are ambiguous, underspecified, or only partially ordered. Addressing this limitation may require integrating additional inference mechanisms to recover latent structure from natural language. 

\subsection{From System Distillation to User Decision: The Role of Interactive Validation}
\bl{}There are no significant differences in trace length or plan length between system-produced filtrations and user-corrected filtrations, indicating that the interactive component of \pthree{} (\textit{interactive validation}) serves as\bk{} confirmation and a lightweight refinement step rather than substantial correction. Across both studies, system filtration removed approximately half of the actions in users' initial traces, while user adjustments introduced no significant further change in trace length. In the \struct{} study, user validation also did not change goal achievement outcomes, as user-filtered and abstracted traces showed the same distribution of full, partial, and non-achievement as system-filtered traces. 
However, interactive validation is not without risk---manual interventions can sometimes increase plan length. For example, in the \open{} study, one participant reselected a previously filtered action, increasing the user-filtered plan length relative to the system-filtered plan (from 1 to 2 steps in the trace, and 8 to 10 steps in the generated plan).

\bl{}\textit{\textbf{Implication: interactive validation poses uncertain benefit.}} There is insufficient evidence that interactive validation poses a benefit or detriment to plan quality or length. While this step may be removed for efficiency of the \tool{} pipeline, we advocate preserving it in order to maintain user agency \cite{horvitz1999principles} and system correction when necessary \cite{amershi2019guidelines}.\bk{}

\subsection{Future Implementations of \tool{}}

We designed \tool{} to be agnostic to any particular type of user interface. We reflect on a few concrete deployment contexts below. 


\bl{}
\paragraph{Language Model Integration} There are many opportunities for language model integration within the \tool{} approach. First, an LLM or VLA could translate the user's natural language input into a symbolic trace, which \tool{} could then distill into a minimal, partially ordered set of critical actions before passing it to a task planner for execution \cite{lee2025veriplan, pmlr-v235-kambhampati24a}. Second, rather than relying on the current symbolic filtering pipeline, an LLM could process the user-created trace directly to identify and remove redundant actions, offering greater flexibility in handling incomplete or malformed input. In both cases, \tool{} serves as a human-in-the-loop validation layer that allows users to inspect and correct the system's interpretation of their intent before it is committed to execution, positioning \tool{} not as a replacement for LLM-based planning but as a complementary layer that addresses the gap between what LLMs infer and what users actually intend.\bk{}

\paragraph{On-the-Fly Task Authoring} Beyond our proof-of-concept web implementation, \tool{} holds promise in on-the-fly authoring domains, such as when the user needs a quick and easy way to task robots on their mobile device \bl{}\cite{stegner2024understanding}\bk{}. Rather than spend cognitive effort on meticulously crafting a concise task specification that the user is confident is correct, reflects their preferences, and affords the robot an appropriate degree of autonomy, the user can instead leverage how they would \textit{naturally} express the task through either natural language or a long-form hand-crafted trace \bl{}\cite{puig2018virtualhome}\bk{}. \tool{} would then appropriately guide the user towards refining their task. 


\paragraph{Embodied Service Robots in Public Environments} \tool{} could also be deployed with embodied service robots operating in public environments such as hospitals, hotels, or eldercare facilities, where users interact with robots through natural language. In these contexts, guests, patients, and care recipients are often untrained and inexperienced in interacting with robots \bl{}\cite{huang2016design}. The \tool{} approach holds promise in affording these individuals the ability to express tasks via spoken language commands, which could then be easily translated to a user trace (\eg{} via an LLM \cite{pmlr-v235-kambhampati24a})\bk{}, alleviating the need for users to hand-craft these traces themselves. Users would still have the benefit of being guided by \tool{} to ensure that the robot's behavior ultimately matches their intent.
\paragraph{Collaborative Industrial and Warehouse Robots} \tool{} could also be implemented in collaborative industrial or warehouse robots that work alongside human operators. In these settings, workers frequently need to specify flexible workflows such as picking, sorting, or assembly while responding to changing inventory, layouts, or production demands. \bl{}Through a demonstration-to-trace pipeline \citep[\eg{}][]{senft2021situated}, \tool{} supports trace-based\bk{} task specification without requiring expertise in robot programming. This approach enables workers to focus on operational goals while relying on the system to manage execution details, potentially improving both usability and adaptability in fast-paced industrial environments.

\subsection{Limitations and Future Work}
Our evaluation was conducted online with simulated environments rather than physical robots, limiting ecological validity. Users may behave differently with physical robots in real environments, particularly regarding trust and error tolerance. Future work should validate \tool{} with physical robot deployments, \bl{} and in doing so, revisit the question of efficiency, usability, and practicality. Our current aim, by contrast, is to validate the core interaction structure of the \tool{} pipeline as a necessary prerequisite to such optimization.

Similarly, we evaluated \tool{} exclusively in simulated delivery tasks, meaning that the ability of the approach to generalize to other domains remains to be seen. Different domains may have different action vocabularies, constraints, and expertise levels. 
Participants relatedly expressed frustration with available actions (see Section \ref{sec:feedback}), indicating that predefined vocabularies can constrain expressiveness. Future work should investigate customizable action vocabularies or hybrid approaches that combine predefined actions with natural language extensions.

Furthermore, \tool{} is accompanied by a multi-phase interaction cost. An important open question is for whom and under what conditions this cost of \tool{} is justified. We argue that \tool{} is most valuable in pooled-interdependent \cite{zhao2020task}, dynamic environments where plan robustness is highly affected by changes in the environment, and where mistakes are costly. Future work should empirically characterize this trade-off and establish clearer guidelines for when \tool{} should be used. \bk{} 

\bl{}Other limitations exist with our use of a symbolic filtering approach. Although our symbolic pipeline provides transparency, interpretability, and determinism, filtration may produce critical actions as false positives if the underlying planning model mischaracterizes which actions are necessary to achieve the task or fails to capture user-specific preferences and constraints. Furthermore, our symbolic approach does not detect or correct user mistakes. Mistakes may occur when user traces contain mistakenly inserted actions or action parameters, upon which the algorithm may silently propagate the error into the filtered trace rather than flagging it for user review.  We therefore advocate for a neuro-symbolic approach as a future direction, leveraging both the flexibility of LLMs and the hard guarantees of symbolic task planners. Specifically, we envision LLMs performing automated filtration and mistake correction, while leveraging a symbolic approach for planning.\bk{}

Additionally, our binary goal achievement metric does not capture partial correctness or goal quality. Future work should include subjective metrics for the degree to which task outcomes align with user intent and preferences, in addition to support for expressing preferences beyond goal specification (as P23S and P38O indicated with their ordering and delivery preferences). We observed substantial individual variation in autonomy interpretation and interface usage, but did not systematically investigate how user characteristics (expertise, trust, cognitive style) moderate these perceptions.

Finally, our operationalization and evaluation of \tool{} in our web interface are limited in scope. This interface captures only single-session usage, whereas longitudinal studies with more ubiquitous hardware (\eg{} mobile phones) would better reveal how specifications evolve with experience and whether users develop more efficient communication strategies. Our tasks involved only a few goals with relatively simple sequences, so future work should investigate how \tool{} scales to complex multi-step tasks with many goals, intricate dependencies, and conditional logic.



\bl{}\section{Conclusion}

We present \tool{}, an approach to producing minimally sufficient robot task specifications from raw human intent. \tool{} relies on several steps: (1-2) \textit{specifying} a task in the form of natural language and/or a sequence of steps in end-user programming fashion, (3) \textit{filtering} the critical actions from the input, (4) \textit{abstracting} critical actions to high-level goals and (5) \textit{relaxing} the order in which actions and goals must be achieved. Our findings indicate that \tool{} reduces the user's task specification towards a more minimal and robust representation of task requirements.\bk{}


\begin{acks}
This research was supported by George Mason University.
\end{acks}

\bibliographystyle{ACM-Reference-Format}




\clearpage
\appendix 
\section{Backend Implementation}
\label{sec:appendix-backend}


The \pthree{} backend is implemented on top of the ISL parser~\cite{porfirio2025interaction}, which represents robot programs as traces of actions. ISL supports filtering traces to their critical actions via Algorithm~\ref{alg:trace-filtering}, which iteratively removes redundant actions from a sequential trace $T$ by exploiting the observation that actions immediately preceding a critical action are often implied by that action's goal. For each critical action $a_c$, the algorithm extracts its positive effects as a goal set $G$ and invokes an optimal classical planner from the current world state $\sigma$ to $G$, yielding a reference plan $\pi$. It then compares the preceding segment of $T$ against $\pi$ in reverse order, identifying the longest suffix of the original segment that appears as a subsequence of $\pi$; these actions are redundant because the planner would reproduce them when satisfying $G$ from $\sigma$. After removing the flagged actions, the world state is advanced by simulating $\pi$ if any removals occurred. The outer loop repeats until a fixed point is reached---\ie{} no further actions can be removed---yielding a filtered trace that retains only the critical actions not recoverable by re-planning between critical actions.

\begin{algorithm}[t]
\caption{\textsc{Filter}($T$)}
\label{alg:trace-filtering}

\KwIn{Sequential trace $T = \langle a_1, a_2, \ldots, a_n \rangle$}
\KwOut{Filtered trace containing only critical actions}

\If{$|T| < 2$}{
    \Return{$T$}\;
}

$T_{\mathrm{prev}} \gets \emptyset$\;

\While{$T \neq T_{\mathrm{prev}}$}{
    $T_{\mathrm{prev}} \gets T$\;
    $\sigma \gets \sigma_0$\;
    \tcp{Initial world state}

    \For{$c \gets 2$ \KwTo $|T|$}{
        $G \gets \textsc{PositiveEffects}(a_c)$\;
        $\pi \gets \textsc{Plan}(\sigma, G)$\;
        \tcp{Optimal plan from $\sigma$ to $G$}

        $\mathcal{O} \gets \textsc{Reverse}(\langle a_1, \ldots, a_{c-1} \rangle)$\;
        $\hat{\pi} \gets \textsc{Reverse}(\pi)$\;

        \tcp{Find redundant prefix of $\mathcal{O}$ via subsequence matching}

        $\mathcal{R} \gets \langle\rangle$; $j \gets 1$\;

        \For{$i \gets 1$ \KwTo $|\mathcal{O}|$}{
            \While{$j \leq |\hat{\pi}|$ \textbf{and} $\mathcal{O}[i] \neq \hat{\pi}[j]$}{
                $j \gets j + 1$\;
            }

            \If{$j > |\hat{\pi}|$}{
                \textbf{break}\;
            }

            append $\mathcal{O}[i]$ to $\mathcal{R}$\;
            $j \gets j + 1$\;
        }

        $T \gets T \setminus \mathcal{R}$\;
        \tcp{Remove redundant actions}

        \If{$\mathcal{R} \neq \emptyset$}{
            $\sigma \gets \textsc{Simulate}(\pi, \sigma)$\;
        }
    }
}

\Return{$T$}\;

\end{algorithm}



\section{Automated Detection of Lexical Features}\label{sec:appendix-reg}

This appendix documents the regular expression–based approach used to identify lexical features in participant natural language input. These cues were used as part of an automated qualitative analysis to uncover instances where participants expressed ordering, conditional reasoning, step-like, or grouping action descriptions.

The analysis focused on four categories: Sequences, Grouping, Conditionals, and Steps. Importantly, the presence of a match indicates the use of a linguistic cue associated with structure, rather than a definitive claim about the participant’s intent or the formal correctness of the described procedure.

\subsection{Regular Expressions for \textit{Sequences}}
Sequences capture lexical cues that suggest temporal ordering or progression of actions or events. Regex patterns that were used to detect sequences include:

\begin{itemize}
\item [] \texttt{[tT]hen}
\item [] \texttt{[fF]inally}
\item [] \texttt{[nN]ext}
\item [] \texttt{[aA]fterwards}
\item [] \texttt{in that order}
\item [] \texttt{[aA]fter}
\item [] \texttt{[fF]irst}
\item [] \texttt{[fF]ollowed by}
\end{itemize}

\subsection{Regular Expressions for \textit{Steps}}
Step indicators capture language suggesting the addition of individual actions within an ongoing process, often without explicit temporal ordering. Regex patterns that were used to detect steps include the same regex patterns for Sequences in addition to:

\begin{itemize}
\item [] \texttt{and( also)? VERB}
\item [] \texttt{([.?!,] )}
\item [] also
\end{itemize}

Note that VERB denotes a placeholder for verb-matching logic implemented elsewhere in the analysis pipeline.

\subsection{Regular Expressions for \textit{Conditionals}}
Conditional indicators capture language that expresses contingency, branching, or dependence of an action on a particular condition. Regex patterns that were used to detect conditionals include:

\begin{itemize}
\item [] \texttt{[iI]f (?!you could)}
\item [] \texttt{in case}
\end{itemize}

\bl{}
\subsection{Regular Expressions for \textit{Grouping}}
Grouping captures lexical cues that suggest actions or events may be treated as belonging together, rather than as strictly ordered steps. Regex patterns that were used to detect grouping include:

\begin{itemize}
\item [] \texttt{[aA]nd}
\item [] \texttt{[bB]oth}
\item [] \texttt{[aA]ll}
\item [] \texttt{[tT]ogether}
\item [] \texttt{\&}
\item [] \texttt{[aA]s well}
\item [] \texttt{[aA]lso}
\item [] \texttt{[aA]t the same time}
\end{itemize}
\bk{}


\end{document}